# A Descriptive Model of Robot Team and the Dynamic Evolution of Robot Team Cooperation

**Shu-qin Li** [1,2], **Lan Shuai** [1], **Xian-yi Cheng** [1], **Zhen-min Tang** [1] & **Jing-yu Yang** [1]
[1] Department of Computer Science of
Nanjing University of Science and Technology, Nanjing, P.R. China
[2] Department of Computer Science of
Beijing Information Science &Technology University,Beijing, P.R. China
lishuqin_de@126.com

*Abstract: At present, the research on robot team cooperation is still in qualitative analysis phase and lacks the description model that can quantitatively describe the dynamical evolution of team cooperative relationships with constantly changeable task demand in Multi-robot field. First this paper whole and static describes organization model HWROM of robot team, then uses Markov course and Bayesian theorem for reference, dynamical describes the team cooperative relationships building. Finally from cooperative entity layer, ability layer and relative layer we research team formation and cooperative mechanism, and discuss how to optimize relative action sets during the evolution. The dynamic evolution model of robot team and cooperative relationships between robot teams proposed and described in this paper can not only generalize the robot team as a whole, but also depict the dynamic evolving process quantitatively. Users can also make the prediction of the cooperative relationship and the action of the robot team encountering new demands based on this model.*

## 1. Introduction

At present, the research of multi-robot has been one of the most popular topics in the robotics realm. In a dynamic, complicated and unknown environment, it is challenging to organize and control robots to finish a task that individual robot hardly do. To make the multi-robot system operate effectively, we should organize it and control it with proper mechanisms. The issue on how to organize and control the multi-robot system has been extensively studied in [Wei-Min Shen & Behnam Salemi 2002] [Torbjorn S.Dahl Maja J. Mataric etc 2002][Tang zhen-min 2002]. But most of these works only focus on individual behavior, cooperation between individuals and between individual and groups, with few works on the cooperation in teams. In the same time, the description methods or models are also very rare, that can qualitatively describe how the robot teams are organized in the changing dynamic environment.
From the social means of topology, structure and organization, this paper studies the multi-robot organization formation and running mechanisms. It presents and describes in detail a Hierarchical-web Recursive Organization Model (HWROM). Based on Markov processing and evolutionary computing, this paper proposed a dynamic evolutional model on robot team cooperation, in order to give a quantitative description on the dynamic evolving process of robot team organization and relationships. Composition of the robot team is described in section 2, team cooperation of the robot teams is discussed in section 3, dynamic evolution model of the cooperation between robot teams is proposed and discussed in section 4.

## 2. General description on robot teams

Through the evolution and development, human society has been developed into a local and hierarchical structure. The basis of the society is individuals with various backgrounds, abilities, knowledge levels and specialties. The behavior and efficiency of the local team decided by different individuals, cultural backgrounds, technology- levels and different management mechanisms. This character, in some sense, provides the organization forms of Multi-robot group bionic evidence.

*Definition 1*
Automatic moving robot: automatic moving robot is a robot that can sense the state of itself and the local environment through the sensors, bringing limited communication equipment, moving toward the goals automatically in the environments with obstacles (static or moving) and finishing certain works. The robot can be made of electromechanical devices (action unit, executing unit, sensing unit and control units), computer and information processing system (intelligent decision-making, action control, moving control, environment sensing and communicating mechanics). We use R to represent automatic moving robot. And we call robot as individual robot.

*Definition 2*
Robot team: Robot team is an organic group with tight connections, organized temporarily by individual robot based on tasks, can be seen as an autonomous domain. The cooperation of the robots in the autonomous domain is called inter-layer cooperation. And the cooperation between the autonomous domains, tight or loose, is called out-layer cooperation. The robot acts as the center of the organization with stronger communicating and organizing ability is called the group leader. The computing actions and a coordinating wills defined by the group leader represent the actions and wills of this group. The number of the robots in the robot teams is changing dynamically; any individual robot can join or quit dynamically.

*Definition 3*
Robot society: the robot society is organized by all of the individual robots and robot groups that have connections, which is a large group with organization.
The top: the topology of cooperation organization of the robot team is shown in Fig.1. In considering the special status of the ultimate user in the actions of the robot group, we list it as a single node in the top left in Fig.1. This model not only represents that the multi-robot system organization has the static and hierarchical character, but also embodies the dynamic character of the forming and cooperating process of the robot team.

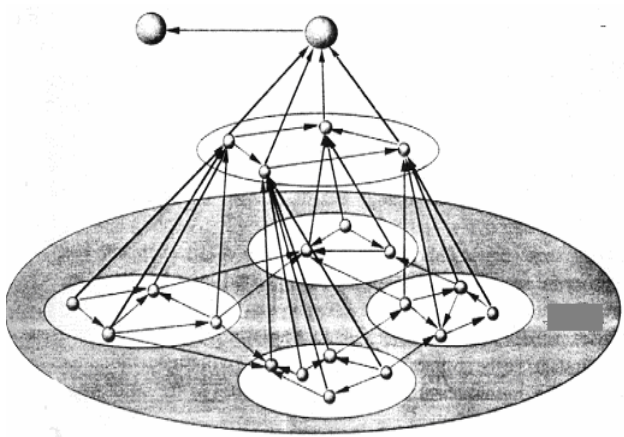

Fig.1 Sketch map of organization topology in multi-robot system cooperation

During the cooperation, the multi-robot team changes with the environment and the demands of tasks. In the same time, the organization of the robot team can change in the same level and also between different levels. Besides, the robot members in the same autonomous domain can coordinates based on the grounds, and form the organization and architecture that is adaptive to the environment. In sum, the robot team is a kind of dynamically, which can keep a dynamic balance.

**3. A formal description of the cooperative relationship in the robot team**

*3.1 A formal description of the robot team*
The robot team is a problem-solving machine of loose coupling networks, which are beyond the ability of individual. Each robot in the network is an independent problems solving machine. Because none of these problem-solving machines has enough specialties, resource and information to solve the question, and different problem solving machines have different specialties to solve different aspects of the question; so to cooperate is a necessity.

*Definition 4*
The individual robots and team leaders in different levels are unified called cooperate robots, abbreviation CR. Cooperative robots have to have certain abilities and make one's own income maximization. It uses a unified frame to describe CR:

CR=<id_cr,Capability,Resource,,Interface>

1. *id_cr*, a identifier of the cooperative robot.
2. *Capability*, the ability of CR.
To a moving robot its ability mainly includes moving ability, action ability, sensible ability, communication ability, organization ability and learning ability, etc.
3. *Resource*, the robot has to need things for finishing his job.
4. Interface, interface way between the cooperative robots.
Definition 5. The multi-robot team structure formed with finite cooperative robots is a hierarchical-web structure.
*Rorg = < {CR}, Re >*
1. *{CR }*, a set of the cooperative robot who carry out certain work in team .
2. *Re*, a dualistic relation set between CR robots, including the control relation structure in the vertical level and the cooperation relation in the horizontal direction.
*Re ⊆ {CR}×{CR}*

*Definition 6.*
A hierarchical-web multi robot organization structure is a recursive structure.

$ROS^{(i,j)} = \{ROS_0^{(i,j)}, ROS_1^{(i,j)},..., ROS_n^{(i,j)}\}$
$ROS_k^{(i,j)} =_{def} < id\_ros, id\_robot, ROS^{(i+1,l)}, T_k^{(i,j)}, CA_k^{(i,j)}$
$C_k^{(i,j)}, R_k^{(i,j)},, EU_k^{(\overline{i},j)} >$

1. *id_ros*, a marker of team structure.
2. *id_robot*, a marker of robot who bears sub-team, namely the robot Team Leader by former identify. Before the robot social organization structure form, the marker is null.
3. $ROS^{i,j}$——a sub-team which lies the *i*th level and the *j*th position in multi-robot team structure. $ROS_k^{(i,j)}$ is the *k*th element in (i,j)'s sub-team structure and maybe also a sub-team structure. $ROS_0^{(i,j)}$ is the leader of the sub-team structure. On the one hand $ROS_0^{(i,j)}$ carries through coordinates with others, responsible division complicated task and team analysis of the result, on the other hand it also does some work that can be finished. There are several sub-teams in a multi-robot society. While *i = 0* it means the highest level of robot society. While $ROS^{(i,j)} = \varphi$ and *i ≠0*, it means a minimum unit of the team structure, namely individual robot.

$$R = < id\_ros, \psi, \varphi, T^{(role)}, C^{(role)}, R^{(role)}, EU^{(role)} >$$

id_ros is a individual robot marker. $\psi$ is a null robot identifier and $\phi$ is the null set.

4. $T_k^{(i,j)}$, the goal set of corresponding cooperative robot. Division $T_k^{(i,j)}$ as

$$T_k^{(i,j)} = \{ t_1^{(i+1,j)},..., t_l^{(i+1,j)},...,t_n^{(i+1,j)} \}$$

$t_l^{(i+1,j)}$ is a sub-goal of sub-team and it can be broken down further according to the request of the task complexity.

5. $C_k^{(i,j)}$, a relationship set which $ROS_k^{(i,j)}$ cooperate and restraint with corresponding cooperative robots in the same level.
6. $R_k^{(i,j)}$, the behavior norm (or behavior rule) set of the robot team organization and distribution rules. All robots in the corresponding robot team must observe.
7. $EU_k^{(i,j)}$, benefit of cooperative result corresponding cooperative robot. It gets from multi-robot teams on or from external environment after cooperative robots have completed their goal tasks.

There are a lot of the cooperative behaviors that refer to the sets of the cooperation relationship and restraint relationship existing between the cooperative robots. These cooperation behaviors form a dynamical net, called cooperative relationship web of multi-robots.

*3.2 A description based on Markov processing cooperative relationships*

The dynamic evolution process of the robot team cooperative relationship can be seen as the process of robot team changing with the environment and the changing demands of tasks, and the adjustment process of the goals, entities and the cooperative relationship between them. The model of the cooperative goals can be embodied by the changing goals in the environment. The pith of the description of the robot cooperative relationships is to describe whether robots can form cooperative relationships under the new demands of tasks in the environmental tasks. So we use finite Markov chain in transition matrix to describe whether the entities cooperate under new demands of tasks. For any two entities, they are always in one of two states, in cooperation or not. If the cooperative relationship exists, then we set the state as 1; otherwise, the state is 0. If the cooperative relationship exists in time t0, then when confronting new environment demands at t1, the relationship has the probability of P11 to still exist, while there is a probability of P10 that the relationship dies. Here, $0 \le P11, P10 \le 1$ and $P11 + P10 = 1$. If there is no cooperative relationship exists in time t0, then when confronting new environment demands at t1, the probability of having the cooperative relationships is p01, while there is a probability of P00 that the relationship is still not exist. Here, $0 \le P00, P01 \le 1$ and $P00 + P01 = 1$. Then we can use the following transition matrix P to describe the cooperative relationships between any robots.

|   | 1   | 0   |
|---|-----|-----|
| 1 | P11 | P10 |
| 0 | P01 | P00 |

Table 1. The probability of the cooperative relationships

The element of the transcription matrix is derived from Bayesian theorem.
We suppose whether or not the cooperative entities cooperate with others is an independent event. And the action set of cooperative relationship between entity $CR_i$ and the relating entity $CR_j$ when confronting new environment demands is set to

$$A_i \quad a_{i1} \quad a_{i2} \ldots \ldots a_{in-1}$$

$a_{ij}$ represents a certain action of constructing a cooperative relationship between $CR_i$ and relating $CR_j$. The priori probability of $CR_i$ choosing to cooperate or not to cooperate with $CR_j$ is $P_{ij}(c_1)$ and $P_{ij}(c_0)$ respectively. Then $0 \le P_{ij}(c_1) \le 1$ ∏ $0 \le P_{ij}(c_0) \le 1$ ∏ and $P_{ij}(c_1) + P_{ij}(c_0) = 1$
When there is a cooperative relationship between $CR_i$ and $CR_j$, then in confronting new demands of tasks, there is a probability of $P(c_1 | a_{ij})$ to observe the action $a_{ij}$. When there is no cooperative relationship between $CR_i$ and $CR_j$, then in confronting new demands of tasks, there is a probability of $P_{ij}(c_0 | a_{ij})$ to observe the action $a_{ij}$. Obviously,

$0 \le P(c_1 | a_{ij})$ ∏ $P(c_0 | a_{ij}) \le 1$ ∏ $P(c_0 | a_{ij}) + P(c_1 | a_{ij}) = 1$

Then the posteriori probability for $CR_i$ taking action $a_{ij}$ is

$$P(a_{ij}) = P(a_{ij} | c_1) \; P_{ij}(c_1) + P(a_{ij} | c_0) \; P_{ij}(c_0)$$

With Bayesian formula, we can get

$$P(c_1 | a_{ij}) = \frac{P(a_{ij} | c_1) \bullet P_{ij}(c_1)}{P_{ij}(a_{ij})}$$

In a similar way,

$$P(c_0 | a_{ij}) = \frac{P(a_{ij} | c_0) \bullet P_{ij}(c_0)}{P_{ij}(a_{ij})}$$

So the elements in the transition matrix are

$$P11 = \sum_{i,j=1}^{V} P(a_{ij} | c_1) \, P(c_1 | a_{ij})$$

P10=1-P11

$$P01 = \sum_{i,j=1}^{V} P(a_{ij} | c_0) \, P(c_0 | a_{ij})$$

P00=1-P01

After the former transitions, and the synthesis of the cooperation of all the cooperative entities, we can get the description of whether or not the cooperative relationship between the entities really changing dynamically with the environment demands.

**4. Dynamic evolution model of the cooperation between robot teams**

The dynamic evolution of the cooperative relationship between robot teams is an adaptive means to solve questions of environment and demands of tasks. According to the adaptation to the solving environment, there are two ways to form the team. One is to design the team structure in advance, based on the question solving demands and various constrains. The team structure does not change in the whole problem solving process. And to solve interaction between robots, the task segregation problem and the load balance problems is relatively simple because of the static robot team structure. The shortcoming is that it can not solve problems well in the open environment or when goals are changing during the problem solving process. And the problem solving ability will decrease or even inaccessible when environment is changing. Another way to organize the robot team is to change the team structure after it has formed by environment and the demands of the problem to be solved. It is called the dynamic evolution of the team cooperative relationship.

Because the team cooperation system is an open system, environment restrictions and the goal of problem solving will change continuously. So the team cooperative relationship with dynamic evolution ability should adapt to environment better, with better problem solving efficiency and lower computational complexity.

Dynamic evolution of the cooperative relationship between robot teams can be recognized by the following characters: temporary cooperation between teams with same goals, in order to complete complex tasks and share risks and interests. Each robot team focus on the core ability of the team itself and take on the strategies to cooperate with outside teams. The dynamic evolution realization of the cooperative relationship between the robot teams does not mean to organize one more layer beyond the team structure, but to decrease the cost of forming a team by deepen inner cooperation and the interacting system between teams.

Dynamic evolution of the cooperative relationship between robot teams is formed by more than two teams cooperating in a limit period of time to benefit each other, get maximum benefits and sense the unknown environment most quickly with the minimum costs. It optimizes the team and dies with the finishing of the common goal. Fig.2 gives a dynamic evolution model of the cooperative relationship between teams, in which the cooperative relationship between teams plays an important role.

This model consists of three layers: cooperative entity layer, ability layer and the relational layer. Robots in the cooperative entity layer have the ability to finish assigned tasks in a given time with limited recourses and maximize its benefits. Robots in the ability layer treat "objects" with strong operability as core. In fact, the computation from the Ce layer to Ab layer is an object-oriented design process. Team is the basic unit of relational layer and it can make each team exert its own advantages to finish the common task quickly and efficiently between different teams.

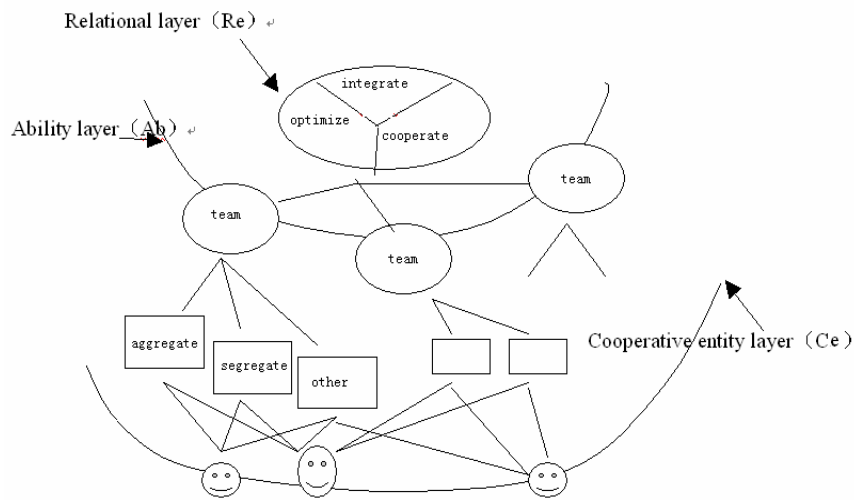

Fig. 2. Dynamic evolution model of the cooperative relationship between teams

Integration, optimization and cooperation are cores of this layer. The dynamic evolution of this cooperative relationship model between teams is finished in the relational layer. Major factors that can effect the cooperative relationship formation between teams are the types of the cooperative entities under new environment demands, which are called payoff function. Suppose the actions behaved in the environment tasks of the robot team are connect to each other, then how to define the best action set based on the payoff function when cooperative entities confronting new environment demands becomes a necessary decision-making problem that must be solved during the dynamic evolution process of the cooperation between robot teams. The basis of the decision-making can be divided into two kinds—to maximize payoff function of the whole robot team and to maximize payoff function of members in robot team.

Suppose new-formed robot team is formed by n cooperative entities, and each entity can build a cooperative relationship with other n-1 entities. We set $a_{ij}$ as the action relating with the formation of cooperative relationship during the interaction of robot $CR_i$ and $CR_j$ under new demands of tasks. $A_i \Pi \Pi a_{i1} \Pi a_{i2} \Pi a_{in-1} \Pi$ represent the action vector of cooperative relationship when robot $CR_i$ interacts with other n-1 entities. We set $u_{ij}(c_1, a_{ij})$ as the payoff function of $CR_i$ cooperating with $CR_j$ when $CR_i$ takes action $a_{ij}$ under new demands of tasks, and $u_{ij}(c_0, a_{ij})$ is set as the payoff function of $CR_i$ not cooperating with $CR_j$ when $CR_i$ takes action $a_{ij}$ under new demands of tasks. $P_{ij}(c_1 | a_{ij})$ is the probability of $CR_i$ choosing to cooperate with $CR_j$ when $CR_i$ takes action $a_{ij}$, and $P_{ij}(c_0 | a_{ij})$ is the probability of $CR_i$ not choosing to cooperate with $CR_j$ when $CR_i$ takes action $a_{ij}$. The whole payoff function of the robot team when taking action set $A_i$ under new demands of tasks is:

$$C_i(A_i) = \sum_{j=1}^{n-1} P_{ij}(c_1/a_{ij}) \; u_{ij}(c_1, a_{ij}) + \sum_{j=1}^{n-1} P_{ij}(c_0/a_{ij}) \; u_{ij}(c_0, a_{ij})$$

Considering the maximization of the robot team member payoff function and the whole robot team, $A^*_i$ should meet the following two demands:

$$C(A^*_i) = \max\{C_i(A_i)\} \text{ and } C(A^*_i) = \max\{\sum_{i=1}^{n} C_i(A_i)\}$$

Here the evolving question of the team formation is converted to the combination optimization question to seek the action set $A^*_i$ when $C_i$ is maximized. The search space of the optimization question is a discrete finite mathematical structure that can be represented by binary coding, to use evolutionary computation to analysis the evolving process of the team cooperation.

## 5. Conclusion

A good robot organization model should be able to improve the efficiency of the system, reduce the complexity of robot interactions, and detract the difficulty of computation. From the sociology aspect of topology, structure and organization, this paper first studies the multi-robot organization formation. It presents and describes in detail a Hierarchical-web Recursive Organization Model. It defines individual robots and team leaders in different levels as the same structure by the united framework and describes the organization model by the recursive structure, so it is helpful to represent the formation of various organizations in the dynamic surroundings.

Second this paper aims at the situation that most of the research on the cooperation between robot teams is still on the stage of quantitatively analysis, with few works on descriptions and forecasts on the dynamic evolution process of cooperation with the changing demands of tasks. We proposed a cooperative mechanism from the cooperative entity layer and ability layer to the relational layer of the robot teams, using Markov process and the evolutionary computation. We also described a dynamic evolving model of the cooperative relationship between robot teams, which is novel now. This model can not only describe the cooperative relationship status between robot teams, but also using finite Markov chain and evolutionary computation to describe whether the dynamic evolution process really forms in the cooperation qualitatively. With this model, users can both handle the cooperative situation globally and dynamically and predict the new cooperative relationship and the actions of robot teams when they confronting new demands of tasks.

## 6. References

Wei-Min Shen & Behnam Salemi(2002).*Distributed and Dynamic Task Reallocation in Robot Organization*,In Proceedings of the 2002 IEEE international Conference on Robotics &Automation ,Washington DC, May 2002,1019-1024

Torbjorn S.Dahl Maja J. Mataric etc.(2002).*Adaptive spatio-temporal organization in groups of robots,* Proceedings of the 2002 IEEE/RSJ international Conference on Intelligent Roboticsand systems ,EPFL Lausanne,Switzerland, October 2002,1044-1049

Tang zhen-min(2002). *Research on Essential Techniques for Mobile Intelligent robot and robot team*, Ph D Dissertation, 2002

Zhang Yan & Shi Mei-lin(2003), *A Description Model for virtual Enterprise Cooperative Relationships Evolution.*Computer Integrated Manufacturing Systems. *Vol.9 No.11 ,966-971*

Li shu-qin,Tang zhen-min etc. *(2004). A self-organization model and formation algorithms of multiple mobile robot.*Application Researchof computer,Vol.21,No.1 2004